\documentclass[10pt,twocolumn,letterpaper]{article}

\usepackage{cvpr}
\usepackage{times}
\usepackage{epsfig}
\usepackage{graphicx}
\usepackage{amsmath}
\usepackage{amssymb}
\usepackage{enumerate}
\usepackage{bm}
\usepackage{threeparttable}
\usepackage{booktabs}
\usepackage{multirow}

\usepackage[breaklinks=true,bookmarks=false]{hyperref}

\cvprfinalcopy 

\def\cvprPaperID{1909} 

\ifcvprfinal\fi
\begin{document}

\title{Collaborative Attention Network for Person Re-identification}

\author{
Wenpeng Li\textsuperscript{\rm 1}, Yongli Sun\textsuperscript{\rm 1}, Jinjun Wang\textsuperscript{\rm 2, 1}\footnotemark[1], Han Xu\textsuperscript{\rm 1}, Xiangru Yang\textsuperscript{\rm 1}, Long Cui\textsuperscript{\rm 1}\\
\textsuperscript{\rm 1}Deep North Inc., \textsuperscript{\rm 2}Xi’an Jiao Tong University\\
{\tt\small \textsuperscript{\rm 1}\{wpli,ylsun,hxu,xryang,lcui\}@deepnorth.cn, \textsuperscript{\rm 2}jinjun@mail.xjtu.edu.cn}
}

\maketitle
\ifcvprfinal\thispagestyle{empty}
\renewcommand{\thefootnote}{\fnsymbol{footnote}}
\footnotetext[1]{Corresponding author, jinjun@mail.xjtu.edu.cn}
\fi

\begin{abstract}
Jointly utilizing global and local features to enhance model performance is becoming a popular approach for the person re-identification (ReID) problem, because previous works using global features alone have very limited capacity at extracting discriminative local patterns in the obtained feature representation.
Some existing works that attempt to collect local patterns explicitly slice the global feature into several local pieces in a handcrafted way.
By adopting the slicing operation, models can achieve relatively high accuracy but we argue that it cannot take full advantage of partial feature slices.
In this paper, we show that by combining neighbourhood local features, we can further improve the final feature representation for Re-ID.
Specifically, we first separate the global feature into multiple local slices at different scale with a proposed multi-branch structure.
Then we introduce an end-to-end Collaborative Attention Network (CAN) in which the features from adjacent slices are comnbined.
In this way, the combination keeps the intrinsic relation between adjacent features across local regions and scales, without losing information by partitioning the global features.
Experiment results on several widely-used public datasets prove that our proposed method outperforms many existing state-of-the-art methods.
\end{abstract}

\begin{figure}[htb]
\centering
\includegraphics[width=0.8\columnwidth]{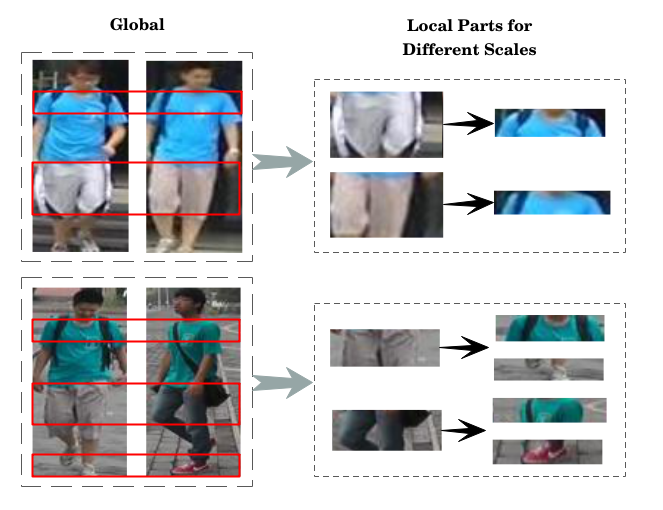}
\caption{These two pairs of bounding box images are from Market-1501 dataset~\cite{zheng2015market}. On the left column are images of various persons with similar clothing and on the right column are local parts partitioned from images by different scales. Models which solely extract global features cannot make a distinction between different persons for each pair. In the upper pair, persons can hardly be distinguished by local features in smaller scale while local parts with larger scale play a more important role in this example. In the lower pair, the colors of shoes which are local details in smaller scale are essential in this case.}
\label{fig1}
\end{figure}

\section{Introduction}

The person re-identification (Re-ID) problem is one of the most significant yet challenging tasks in the computer vision area.
The input data for person Re-ID system are mostly the full body bounding boxes of pedestrian detected from multiple surveillance cameras.
The person Re-ID system then helps to retrieve a certain person among all the bounding box images by figuring out under which cameras the person appears and re-appears.
Comparing with a typical face recognition system where facial images are usually obtained from a constrained environment, person Re-ID is more applicable in real-world scenario, whereas factors such as occlusion, posture change and illumination variation among different cameras further complicate this problem.

In recent years, deep neural networks have been proven effective at extracting discriminative features for the images classification problem~\cite{Jia2009ImageNet, he2016deep, HuSqueeze}, and are therefore widely used as base models of person Re-ID approaches.
For instance, previous works~\cite{fu2018towards, luo2019bags} proposed standard baselines in which ResNet~\cite{he2016deep} was used to extract the appearance feature of the full body image.
Since such network model obtains only a global representation, model performance is limited at distinguishing bounding box images with minor difference.
The baseline methods based on~\cite{he2016deep} only obtains a global feature such as the overall color of clothing which is insufficient for person Re-ID.
Some examples are shown in Figure~\ref{fig1}.
Global features for persons with similar clothing are close to each other in feature space and not sufficient for distinguishing different persons in each pair.
Typically, we identify persons not only by general clothing colors but also by local details which cannot be obtained by global feature representations.
To improve the performance of models that extract only the global feature, models are enabled to extract local features which are attached to original global features for more discriminative representation capacity.
For example, ~\cite{rui2014learning} addressed the issue by collecting representative patches from the image, and the trained model show good balance between discriminative power and generalization ability.
For a more explicit part-based feature representation for Re-ID, ~\cite{yao2019deep, zhao2017deeply} utilized firstly a part detector to locate body parts and then extracted features from different local parts.
Although experimental results validate that combining global and local features can improve the model accuracy, the definition of what parts to locate is rather arbitrary and may change dataset by dataset and/or object class by class.

To achieve a more data-driven way, researchers have also investigated the attention mechanism to implicitly infer the importance of different local regions.
For example, in~\cite{ChenMixed} the author proposed the High-Order Attention module that contains high-order statistics of convolutional activations.
At the same time, since the part-based methods can also be viewed as learning the attention among parts, end-to-end method based on learning the importance of parts or part combinations have also been reported.
To give an example, ~\cite{Sun2017Beyond} proposed a part-based Re-ID models where mid-level features were divided into slices with fixed spatial location and size.
Local features from these slices were then arbitrarily fused to derive the final representation.
The method can be viewed as a hard combination mechanism with identical attention for each part, and leads to state-of-the-art performance.
On the other hand, since the global level information will be lost when global feature is separated into too many slices, defining the proper parts or part combinations becomes challenging and may easily overfit.
Our experiments on \textit{number of parts} hyperparameter show that model accuracy will descend when increasing the granularity of slicing feature maps.

In this paper, we propose the Collaborative Attention Network to further improve over existing part-based Re-ID feature models.
The idea is to combine features from adjacent slices to keep the intrinsic relation between adjacent features across neighbourhood regions in the same scales, without losing information by partitioning the global features, while generating more discriminative features based on local patterns.
The contributions of this paper are two-fold:

\begin{enumerate}[(i)]
\item We address that finer local features are beneficial for more discriminative feature representation and by local feature combination in our method, we can make full use of local features.
\item Local features can refer the relative information from neighbourhood regions by introducing the Collaborative Attention Network, which makes the part-based more effective.
\end{enumerate}

\section{Related Works}

With the massive development of deep learning and neural networks, person Re-ID methods have made great progress from traditional hand-craft feature extraction to deep-learned neural network models.
Release of public person Re-ID dataset~\cite{zheng2015market, ristani2016MTMC, Wei2014DeepReID, Wei2018Person}, researchers can train Re-ID models over a large scale of image data.
Some deep learning methods with tricky strategies to specifically deal with person Re-ID problems are proposed.
Then, partial features are proven to be useful because persons are distinguished between each other by local details along with general global features.
So recently, part-based methods and attention mechanism are becoming popular.

\begin{figure*}[htb]
    \centering
    \includegraphics[width=1.0 \textwidth]{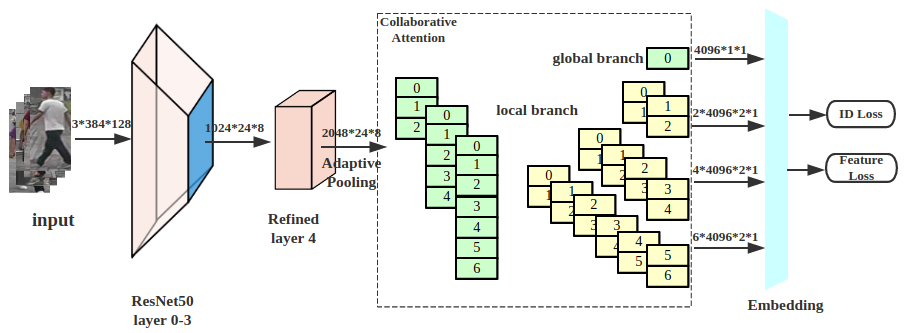} 
    \caption{Pipeline of our proposed Collaborative Attention Network (CAN).
             ResNet50 is used as base model but we refine the Layer 4 in ResNet50 to give out more detailed features.
             Collaborative Attention method is introduced for partial feature branches.
             A multi-loss function is employed to make the feature representation more discriminative.}
    \label{fig2}
\end{figure*}

\subsection{Deep Learning Methods}
Compared to hand-crafed methods, CNN-based deep-learned systems are much more well-performed as long as large scale of data can be achieved.
In~\cite{zheng2016person}, Zheng et al. provided a person Re-ID summary in the early stage, in which some methods fine-tuned from image classificaiton models were mentioned.
Sun et al. proposed the SVDNet in~\cite{Sun2017SVDNet} in which a feature embedding layer was attached after the backbone network to improve the performance.
Since person Re-ID is viewed as a retrieval problem, most of the previous methods only focused on probe-to-gallery (P2G) affinities during testing stage.
In~\cite{shen2018deep}, authors novelly conducted gallery-to-gallery affinities to refine the P2G affinities.
Considering the limitation of pedestrian detectors, a person alignment network was proposed in~\cite{Zheng2017Pedestrian} which can align the pedestrians within bounding boxes and learn the feature descriptor at the same time.
And in~\cite{ZhangDensely}, authors firstly made use of the grained semantics to fix misalignment problem.
For multi-task learning, some of the methods used metric loss along with classification loss to form the multiple loss functions.
Hermans et al. introduced the Triplet Loss into the person Re-ID task in~\cite{Hermans2017In} which improve the model performance by a large margin.
Muti-task learning has become a common stategy when training a person Re-ID model.

\subsection{Methods with Attention Mechanism}
Developed from deeply-learned methods in the early stage and in order to solve misalignment problem, some models which rely on attention mechanisms were proposed.
A content-aware network was proposed in~\cite{li2017learning} to learn discriminative features over full body and local body part.
In~\cite{zhao2017deeply}, Zhao et al. reported that body part extractor inspired by attention models can improve the performance combined with feature calculation.
Li et al. in~\cite{LiHarmonious} addressed the effectiveness of combining attention selection and feature representation in an end-to-end learning process.
Also, the hard regional attention and the soft pixel-wise attention were both used to deal with multiple levels of attention subjects.
Body landmarks and person pose estimation can also be viewed as an attention mechanism.
Some methods used body segmentation network or landmark detectors~\cite{Shelhamer2017Fully, yao2019deep, Insafutdinov2016DeeperCut} as the basis of the main body attention for a model and achieved great progress.
However, since datasets for training such detectors are different with person retrieval datasets, these methods do not perform perfectly.

\subsection{Part-based Methods}
The major contribution of this paper is a more effective part-based model.
We compare our proposed model with existing part-based methods which are useful when learning feature representation for person Re-ID.
Inspired by the attention models, some researchers established some simple but strong and useful part-based methods.
In~\cite{Sun2017Beyond}, the author proposed the Part-Based Convolutional Baseline (PCB) in which a uniform partition on global features was used.
Then these local features were used for classification and this method was verified to be effective yet not to be improved.
Along with PCB model in this paper, the author introduced a refined part pooling (RPP) method.
RPP method is the attention mechanism that can re-assign parts by image blocks instead of uniformly slicing the images.
Inspired by this part-based method, recent work~\cite{Wang2018Learning} put forward the novel Multiple Granularity Network (MGN) which is the state-of-the-art method on person Re-ID benchmarks.
MGN is based on a multi-branch structure where global features and local features are combined together as the final feature representation.
In branches of local features, images are separated into 2 and 3 stripes from original feature maps.
By diversity of granularity, different branches are varied in feature extraction preference.
Combining global features with multi-granularity local features, the feature extractor is more comprehensive.
Our method in this paper is inspired by the multi-branch structure but with finer feature partition and with a more effective collaborative attention method.

\section{Method}

\subsection{Overview}

Collaborative Attention Network (CAN) is a globally-and-locally muti-branch network and the pipeline is shown in Figure~\ref{fig2}.
Generally, the pipeline can be separated into: base feature extractor, Collaborative Attention module, feature embedding layer and the loss function.
In our proposed structure, global features are firstly sliced into partial features by different scales to form multiple branches and then both local and global features are combined to obtain discriminative feature representations.

\subsection{Network Structure}
In this part, the detail of our network structure will be introduced in details including the base model, Collaborative Attention module and the feature embedding layer.

\begin{figure}[htb]
\centering
\includegraphics[width=1\columnwidth, height=4cm]{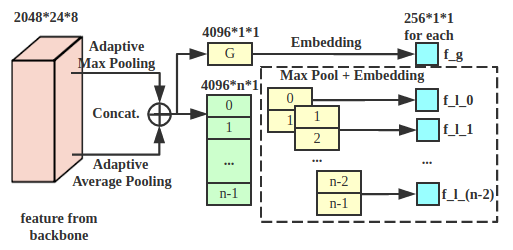}
\caption{Collaborative Attention module for local branch. The feature map from the base model for each local branch is passed into Adaptive Pooling Layer and outcomes the global block and local feature map with specific size. Then, adjacent local blocks are collaboratively used as local features in our proposed part-based method. Global and local features are embedded to form the final feature representation.}
\label{fig3}
\end{figure}

\vspace{8 pt}
\noindent{\bf Base Model for Feature Extraction}. Typically for person Re-ID problem, deep neural networks for image classification are utilized as base models.
In our proposed CAN structure, ResNet50~\cite{he2016deep} is used as the base model since it gives outstanding performance for image classificaiton.
${Layer~0}$ to ${Layer~3}$ in ResNet50 are kept the original forms while ${Layer~4}$ is modified for the sake of following model extension.
Since the global features are evenly separated into several parts after the base feature extraciton, features can be enriched if feature maps with larger size can be obtained from base model.
Under this consideration, we modify the Bottleneck Layer in ${Layer~4}$ of ResNet50 with a stride of 1 for convenience of cutting global features into more pieces.
After modified ${Layer~4}$, 2048-dimensional features with the size of ${24\times8}$ can be obtained.

\vspace{8 pt}
\noindent{\bf Collaborative Attention Module}. The pipeline is separated into one global branch and three local branches from modified ${Layer~4}$.
We also retain the global features in all local branches along with the global branch.
Since we argue that global features can reveal distinct attributes in different branches, neural weights of ${Layer~4}$ are not shared among all branches.
This Collaborative Attention module is shown in Figure~\ref{fig3}.

After we obtain the feature maps from ${Layer~4}$, we need to horizontally slice the feature maps into different parts for each branch.
So firstly Adaptive Pooling is introduced to generate feature maps with different scales.
In Adaptive Pooling Layer, when the output size is fixed and input size is given by the base model, other hyperparameters of pooling module can be infered by the following equations (assuming ${padding}$ in pooling module equals to 0):

\begin{equation}
\left\{
\begin{array}{lr}
padding = 0 \\
stride = \bm{floor}(IS~/~OS) \\
\begin{aligned}
kernel\_size = IS - (OS - 1) * stride \label{eq.1}
\end{aligned}
\end{array}
\right.
\end{equation}

\noindent where ${IS}$ is input feature map size and ${OS}$ represents output feature map size.
From the Figure~\ref{fig3}, we operate both Adaptive Max Pooling and Adaptive Average Pooling and then concatenate the outputs together before passing into next step.
Former experiments have reported the effectiveness of such Adaptive Concatenation Pooling module.
Because some feature information can be lost either in max pooling or in average pooling, we expand the feature dimensions by concatenating max and average pooling output, which can also enrich the feature information.
The size of feature maps are ${3\times1}$, ${5\times1}$ and ${7\times1}$ respectively for each local branch which will be sliced into parts.
As what is shown in Figure~\ref{fig3}, besides the feature map that will be divided into partial features, a global feature block ${G}$ with size of ${1\times1}$ is generated for the reason that for different scales of local features, the global features learning by the neural network could be varied.
These global features from local and global branches are then delivered into embedding layer which will be introduced in the following subsection.
As a result, after the Adaptive Pooling Layer in the network structure, the details of feature dimensions for each branch is shown in Table~\ref{table1}

\begin{table}[t]
\centering
\resizebox{\columnwidth}{!}{
    \begin{threeparttable}
        \begin{tabular}{ccc}
            \toprule
            {\bf{~~~branch~~~}}&{\bf{~~~~~feature map size~~~~~}} & {\bf{~~feature dimension~~}}\cr
            \midrule
            global&${1\times1}$&${2048\times2}$\cr
            part\_3&${3\times1 + 1\times1}$(global)&${2048\times2}$\cr
            part\_5&${5\times1 + 1\times1}$(global)&${2048\times2}$\cr
            part\_7&${7\times1 + 1\times1}$(global)&${2048\times2}$\cr
            \bottomrule
        \end{tabular}
    \caption{Feature map sizes and feature dimensions for each branch after the Adaptive Pooling Layer.}\label{table1}
    \end{threeparttable}
}
\end{table}

After achieving a multi-branch feature structure, Collaborative Attention module is proposed to deal with local features in each branch.
Figure~\ref{fig3} shows how the Collaborative Attention works for ${part\_n}$ branch in which local feature $\mathbf{f_L}$ is in size of ${n\times1}$.
Local features are evenly divided into ${n}$ blocks ${\{\mathbf{f_{L_0}}, \mathbf{f_{L_1}}, ..., \mathbf{f_{L_{(n-1)}}} \}}$ with size of ${1\times1}$ for every block.
As what is mentioned in~\cite{Sun2017Beyond}, such partition operation is a hard attention mechanism in which we make the learning process focus on the separated parts in a manual way.
In our method, we concatenate two adjacent local blocks ${\mathbf{[{f^{T}_{L_k}}, {f^{T}_{L_{(k+1)}}}]}^\mathbf{T}}$ as the Collaborative Attention mechanism.
Then, we use the Max Pooling to downsample the local features to obtain partial features ${\mathbf{f_{LP_k}}}$, which make the local features and auxiliary global features have identical feature map sizes.
With help of Collaborative Attention module, richer information can be included to enhance the learning process without losing spatial neighbourhood features.
This module is applied in each local feature branch and experimental results show that it can give out a better performance comparing with existing part-based methods.

\vspace{8 pt}
\noindent{\bf Embedding Layer}. Global features from Adaptive Pooling and local features from Max Pooling are all have the dimension of ${4096}$, which will be excessively redundant when features are combined to form the final feature representation.
To fix this issue, a ${1\times1}$ convolutional layer is employed as the embedding layer.
Convolutional layer with kernel size of ${1\times1}$ is widely used as feature map encoders while it can also reduce the feature dimensions.
In our method, we reduce the ${4096}$-dim features to the dimension of ${256}$ for each feature maps and the weights are shared among all branches.
The weights set of this layer is ${\mathbf{W^T} = \{\mathbf{w_1^T}, \mathbf{w_2^T}, ... \mathbf{w_m^T}\}}$ where ${m = 256}$ is the final feature dimension for each branch.
The features from branch ${part\_n}$ is shown in the following Equation~\ref{eq.2}:

\begin{equation}
\begin{split}
\mathbf{F_n} &= \{\bm{f_G}, \bm{{f_{LP}}_0}, ..., \bm{{f_{LP}}_{n-1}}\} \\
&= \{\bm{{f_n}_0}, \bm{{f_n}_1}, ..., \bm{{f_n}_{n-1}}\}
\end{split}
\label{eq.2}
\end{equation}

\noindent in which ${n = 3, 5, 7}$ represents different branches where the feature maps are divided into 3, 5 and 7 parts.
Then the Embedding Layer can be expressed by Equation~\ref{eq.3}:

\begin{equation}
\begin{split}
\bm{W^T}\bm{F_n} &= \{\bf{W^T}\bm{f_G}, \bf{W^T}\bm{{f_{LP}}_0}, ..., \bf{W^T}\bm{{f_{LP}}_{n-2}}\} \\
&= \{\bm{W^T}\bm{{f_n}_0}, \bm{W^T}\bm{{f_n}_1}, ..., \bm{W^T}\bm{{f_n}_{n-1}}\} \\
&= \{\bm{{f_i}_0}, \bm{{f_i}_1}..., \bm{{f_i}_{n-1}}\} \\
\label{eq.3}
\end{split}
\end{equation}

\noindent For each local branch, the feature vector is ${\bm{{f_i}_0}}$ is the embedded global feature and the others are embedded local features.
Each feature map ${\bm{{f_i}_k}}$ where ${k = 0, 1, .., (n-1)}$ equals to:

\begin{equation}
\begin{split}
\bm{{f_i}_k} &= \bm{W^T}\bm{{f_n}_k} \\
&= [\bm{w_1^T}\bm{{f_n}_k}, \bm{w_2^T}\bm{{f_n}_k}, ... \bm{w_m^T}\bm{{f_n}_k}] \\
&= [{{f_i}_k}_0, {{f_i}_k}_1, ..., {{f_i}_k}_{m-1}]
\label{eq.4}
\end{split}
\end{equation}

\noindent where ${m = 256}$ is the dimension of every feature map after embedding.
After processed by our proposed Collaborative Attention Network, we obtain four 256-dimensional global feature maps and twelve 256-dimensional local features maps.

\subsection{Loss Function}

The design of loss function is crucial during training process.
Typical strategy of loss function for person ReID task is to combine ID loss and Triplet Loss.
ID loss can make the model perform well when doing for person classification and Triplet Loss is beneficial for obtaining more discriminative features.
Along with ID loss and Triplet Loss, center loss~\cite{luo2019bags, Wen2016A} is also applied in our loss function.
The overall loss function can be formulated as:

\begin{equation}
\mathcal{L} = \mathcal{L}_{CE} + \mathcal{L}_{Trip} + \lambda\mathcal{L}_{C}
\end{equation}

In this multi-loss function, CrossEntropy is commonly employed in classification problem as the ID loss.
After the FC Layer and Softmax Layer, the output vector is probabilities for different classes which can be expressed as ${\bm{q}}$ and the ground truth for this feature is ${\bm{p}}$ which is a one-hot vector.
Then the CrossEntropy can be formulated as:

\begin{equation}
\mathcal{H}(\bm{p}, \bm{q}) = -\sum\limits_{i=1}^k{p_i\log{q_i}}
\end{equation}

\noindent where ${k}$ is the class number.
Minimizing the CrossEntropy can make the predicted probabilities close to the ground truth.

In our training strategy, for a single mini-batch, there are different IDs with several bounding box images for each ID.
So Triplet Loss and center loss are designed to guide the training process with feature distance.
For Triplet Loss ${\mathcal{L}_{Trip}}$ in this equation, we use hard Triplet Loss:

\begin{equation}
\begin{split}
\mathcal{L}_{Trip} = \sum\limits_{i=1}^{N} {\biggl[\alpha + {\left\Vert {\bm{f_i^{a}}-\bm{f_i^{p}}} \right\Vert}_2^{2}} - {\left\Vert {\bm{f_i^{a}}-\bm{f_i^{n}}} \right\Vert}_2^{2}\biggr]_+
\end{split}
\end{equation}

\noindent where $\bm{f_i^{a}}$, $\bm{f_i^{p}}$, $\bm{f_i^{n}}$ are the anchor feature, positive feature and negative feature, respectively. 
In order to improve performance after introducing Triplet Loss, we use the batch hard positive and negative features, which means:

\begin{equation}
\left\{
\begin{array}{lr}
\bm{f_i^{p}} = \mathop{\arg\max}_{f_i^{p}} {{\left\Vert {\bm{f_i^{a}}-\bm{f_i^{p}}} \right\Vert}_2^{2}} \\
\bm{f_i^{n}} = \mathop{\arg\min}_{f_i^{n}} {{\left\Vert {\bm{f_i^{a}}-\bm{f_i^{n}}} \right\Vert}_2^{2}}
\end{array}
\right.
\end{equation}

\noindent Center loss ${\mathcal{L}_C} $is designed to make samples with identical ID close to clustering center of this ID, which is formulated as:

\begin{equation}
\mathcal{L}_C = \frac{1}{2} \sum\limits_{i=1}^{m}{\left\Vert \bm{x}_i - {\bm{c}_y}_i \right\Vert}
\end{equation}

\noindent where ${{\bm{c}_y}_i}$ denotes the feature center for the class of ${y_i}$.




Supervised by the multi-loss function with Cross Entropy Loss, hard Triplet Loss and Center Loss, the training process can be more effective and the model can give out more discriminative features.
The benefits of utilizing such type of multi-loss function can be seen from the experiments result in the next section.

\section{Experiments}
In this section, we report the experimental results on our proposed Collaborative Attention Network.
All the experiments are mainly conducted on the dataset of Market-1501~\cite{zheng2015market} which is the most widely-used image-based person ReID dataset.
Besides Market-1501, in order to verify the robustness of our model, some other mainstream public ReID datasets are used which are DukeMTMC-ReID~\cite{ristani2016MTMC} and CUHK03~\cite{Wei2014DeepReID}.
Firstly public datasets and the benchmarks to measure the model accuracy are introduced.

\subsection{Datasets and Benchmarks}
\textbf{Market-1501} This dataset was built in the summer of 2015 and released by Zheng et al. and it is the most popular and high-quality image-based person ReID dataset.
This dataset was captured by six cameras (five high-resolution cameras and one low-resolution camera) on campus of Tsinghua University.
The dataset was separated into training set and testing set which is furtherly split into gallery and query for testing set.
There are 1,501 person IDs and 12,936 bounding box images in total, in which one certain person was captured by at least two cameras.
Bounding boxes in the query set are manually labeled while in the gallery set were labeled by DPM detector.

\textbf{DukeMTMC-ReID} This is a subset of DukeMTMC dataset and specificly for the person ReID task.
There are 16,522 training images from 702 person IDs, 2,228 query images from another 702 person IDs and 17,661 gallery images from the same person IDs as query set.
Images are sampled from video tracks by every 120 frames and all the videos are captured by eight high-resolution cameras on Duke campus.
To maintain the high quality of bounding boxes, this dataset was labeled by hand instead of using pedestrian detectors.

\textbf{CUHK03} This dataset is stored as MAT format file.
There are 1,467 different person IDs in this dataset collected by 5 pairs of cameras.
The total 13,164 images are varied in image sie and are separated into three groups: manually-labeled images for training, DPM-detected images for training and testing set.
Along with the dataset, the paper mentioned two types of testing protocols.
In order to keep the consistency of conclusions on different datasets, we chose the testing method where testing set is splited into query and gallery.


\subsection{Comparison of the Number of Sliced Parts}
For our part-based method, we uniformly partition the feature maps into different number of slices.
We argue that such slicing operation can theoritically help with obtaining local features much more detailed and discriminative which are beneficial for person ReID tasks.
If too many parts are obtained from global feature maps, however in contrast, such over-detailed local features cannot retain representative attributes, which would conversely deteriorate the model performance.
The following experiments are comparing the model performance when changing the number of how many parts local features are sliced into and the results are shown in Table~\ref{table2}.

\begin{table}[t]
    \centering
    \resizebox{\columnwidth}{!}{
        \begin{threeparttable}
            \begin{tabular}{ccccc}
                \toprule
                {patterns of how to partition}&{mAP}&{rank-1}&{rank-3}&{rank-5}\cr
                \midrule
                part-1, 2, 3&86.6&94.4&97.2&97.9\cr
                part-1, 2, 3, 4&86.4&94.4&97.2&98.0\cr
                part-1, 3, 5&86.1&94.6&97.5&98.1\cr
                part-1, 3, 5, 7&85.4&94.3&97.2&98.0\cr
                part-1, 3, 5, 7, 9&83.9&93.9&96.9&97.9\cr
                \bottomrule
            \end{tabular}
        \caption{Comparison between different levels of how to partiton global features into local features.}\smallskip \label{table2}
        \end{threeparttable}
    }
\end{table}

From Table~\ref{table2}, it can be seen that ${part}$-${1,2,3}$ which means the global features are evenly divided into 2 and 3 parts can achieve the highest accuracy with mAP equals to 86.6\% and rank-1 equals to 94.4\%.
For the experiment of partial features the accuracy is that mAP equals to 86.6\% and rank-1 equals to 94.4\% which is no higher than ${part}$-${1,2,3}$.
Since the part-4 partial features are included in part-2 partial features, it cannot bring in improvement if features are separated into multiple even numbers of parts.
So in the following experiments in this Table, we divide feature maps into 3, 5, 7 and 9 parts and add branches of different partition levels step by step.
It is clear that when gradually adding feature branches, the accuracy of models is descending and for ${part}$-${1,3,5,7,9}$ the accuracy is 83.9\% for mAP and 93.9\% for rank-1.
We can conclude from the experimental results that simply increasing partial granularities cannot contribute to the improvement of models.
In the next section, the effectiveness of our proposed method is proven by experimental results.

\subsection{Effectiveness of Collaborative Attention}

The Collaborative Attention (CA) mechanism is the key part in our proposed method.
Improved from part-based methods, we concatenate neighbouring local features as collaborative local features in our network.
We argue that our method can take advantage of subtle detail features and we could obtain more discriminative feature representations by combining partial features together.
The results of methods with our proposed Collaborative Attention mechanism are shown in Table~\ref{table3}

\begin{table}[t]
    \small
    \centering
    \resizebox{\columnwidth}{!}{
        \begin{threeparttable}
            \begin{tabular}{cccccc}
                \toprule
                {Method}&{mAP}&{rank-1}&{rank-3}&{rank-5}&{rank-10}\cr
                \midrule
                part-1,2,3 with CA&87.4&94.8&97.6&98.4&99.1\cr
                part-1,2,3,4 with CA&86.7&94.5&97.5&98.3&99.1\cr
                part-1,3,5 with CA&87.6&95.0&97.8&98.4&99.2\cr
                part-1,3,5,7 with CA&87.9&95.2&97.5&98.3&99.2\cr
                part-1,3,5,7,9 with CA&87.2&94.8&97.4&98.2&98.9\cr
                \midrule
                part-1,3,5,7 w/o CA&85.4&94.3&97.2&98.0&98.8\cr
                {\bf part-1,3,5,7 with CA}&{\bf 87.9}&{\bf 95.2}&{\bf 97.5}&{\bf 98.3}&{\bf 99.2}\cr
                \bottomrule
            \end{tabular}
        \caption{Comparative experiments of part-based methods without/with Collaborative Attention mechanism.}\smallskip \label{table3}
        \end{threeparttable}
    }
\end{table}

Comparing Table~\ref{table2} and Table~\ref{table3}, the difference between conducted experiments is that Collaborative Attention mechanism is adopted after features are partitioned.
From each partitioning pattern experimental results, we conclude that our proposed CA mechanism can bring in improvement.
In all combinations of how to separate features, ${part}$-${1,3,5,7}$ achieves the best performance in which the CA mechanism raises model accuracy from mAP 85.4\% and rank-1 94.3\% to mAP 87.9\% and rank-1 95.2\%.
In Table~\ref{table2}, part-based methods with higher level of partitioning show no advantage.
After introducing the CA mechanism, more detailed partial features can have positive effects on producing more discriminative feature representations.
But finer segmentation does not always bring a positive impact, and a local stripe that is excessively thin can no longer contain obvious feature infomation.
Even with the CA mechanism, the improvements are limited.
As a result, the performance of ${part}$-${1,3,5,7,9}$ is no better than ${part}$-${1,3,5,7}$.

\subsection{Combination of Loss Functions}

Multi-loss functions are used in our proposed method, in which we combine ID loss with Triplet Loss and Center Loss.
ID Loss is to improve the classificaiton abilities for the model while Triplet Loss and Center Loss can make output features more discriminative.
In the experiments, we compare the results when we progressively introduce CrossEntropy ${\mathcal{L}_{CE}}$, triplet Loss ${\mathcal{L}_{trip}}$ and Center Loss ${\lambda\mathcal{L}_{C}}$.
We also argue that it would be useful if the outputs from both global and local branches are guided by Triplet Loss and Center Loss, unlike the loss function strategy in MGN~\cite{Wang2018Learning} in which only global features play a part in Triplet Loss.
The results for loss-function-related experiments are shown in Table~\ref{table4}

\begin{table}[t]
    \small
    \centering
    \resizebox{\columnwidth}{!}{
        \begin{threeparttable}
            \begin{tabular}{cccccc}
                \toprule
                {Loss Combination}&{mAP}&{rank-1}&{rank-3}&{rank-5}\cr
                \midrule
                CAN+${\mathcal{L}_{CE}}$&87.0&95.0&97.8&{\bf 98.7}\cr
                CAN+${\mathcal{L}_{CE}}$+${\mathcal{L}_{Trip}}$&87.9&95.2&97.5&98.3\cr
                CAN+${\mathcal{L}_{CE}}$+${\mathcal{L}_{Trip}}$+${\mathcal{L}_{C}}$&88.3&95.4&97.6&98.4\cr
                \midrule
                {\bf global \& local to ${\mathcal{L}_{c}}$}&{\bf 89.6}&{\bf 95.7}&{\bf 97.8}&98.6\cr
                \bottomrule
            \end{tabular}
        \caption{Experimental results of progressively adding different loss functions and comparative results between solely global features and twofold (global and local) features in clustering loss function ${\mathcal{L}_{c} = \mathcal{L}_{Trip} + \mathcal{L}_{C}}$. }\smallskip \label{table4}
        \end{threeparttable}
    }
\end{table}

If only ID Loss (CrossEntropy) is employed as loss function, our proposed Collaborative Attention Network (CAN) can achieve the performance mAP/rank-1=87.0\%/95.0\%.
Triplet Loss and Center Loss can enhance features clustering attributes and we name these two loss functions as clustering loss.
We firstly take global features into calculation of clustering loss as what the existing methods do.
After combining Triplet Loss with ID Loss (CrossEntropy), CAN achieves mAP/rank-1=87.9\%/95.2\% and when Center Loss is adopted, the accuracy is improved to mAP/rank-1=88.3\%/95.4\%.
Then we furtherly use features from local branches as input of clustering loss to guide the training process, the model achieves the best performance with mAP 89.6\% and rank-1 95.7\%.
So supervising each local feature by loss functions during training process can enhance the feature performance when testing.

\begin{figure}[t]
\centering
\includegraphics[width=0.9\columnwidth]{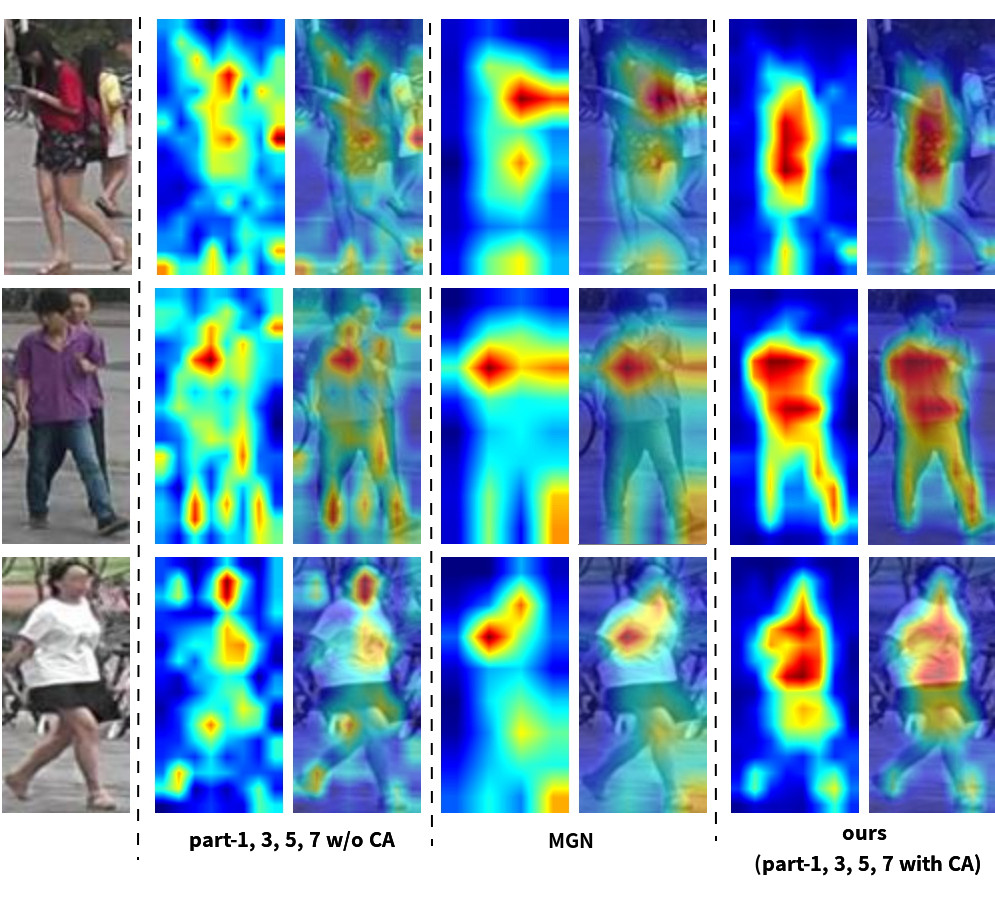}
\caption{Visualization of feature maps before the last pooling layer in the networks of different methods. The left column of feature maps are from the model of part-1,3,5,7 without introducing Collaborative Attention. The middle column is from the output of MGN~\cite{Wang2018Learning} while the right column are the feature maps from our proposed method.}
\label{fig4}
\end{figure}

\subsection{Visualization of the Method}
To directly prove the effectiveness of our method, we visualize and compare the feature maps of different methods in Fig~\ref{fig4}.
The features maps are before the last pooling layer of the network for each method, which can be viewed as the final feature representations.
When we partition the global features into several parts without using Collaborative Attention, the features maps are over-fined, which makes the global feature information is lost.
In the left column, it is clear that the feature maps are merely notable on some local parts.
In the method of MGN, features are splited into two and three parts which are suitable granularities to slicing human body features.
However, the feature maps are over-separated on the basis of the granularities.
To take advantage of the finer local details for the method in the left column, we introduce our proposed Collaborative Attention.
From the right column, we can clearly see that the feature maps are focused on the whole human bodies without losing the neighbourhood feature information for adjacent local parts.
As a result, our method give out the most discriminative feature representations which outperforms other existing methods.

\subsection{Implement Details}

In this section, we introduce the details of how we set up the training strategy for the Collaborative Attention Network.
Since we apply Triplet Loss and Center Loss in our loss function, we set the mini-batch to 32 images in which 8 different person IDs are randomly picked with 4 bounding boxes images for each ID.
All input bounding boxes are resized to ${384\times128}$ with height of ${384}$ and width of ${128}$.
The loss function is a multi-loss function with CrossEntropy, Triplet Loss and Center Loss and the weight of Center Loss in this function is set to 0.0005.
Adam optimizer is utilized with default parameters (${\epsilon = 10^{-8}, \beta_1 = 0.9, \beta_2 = 0.999}$).
The initial learning rate is ${3 \times 10^{-4}}$ and the learning rate is decayed to ${3 \times 10^{-5}}$ at epoch 250, decayed to ${3 \times 10^{-6}}$ at epoch 350 and decayed to ${3 \times 10^{-7}}$ at epoch 450.
The whole training process stops at epoch 600.
We also normalize the features and weights in FC Layer so we use the cosine distance as metrics to measure feature similarity during testing process.


\begin{table}[t]
    \small
    \centering
    \resizebox{\columnwidth}{!}{
        \begin{threeparttable}
            \begin{tabular}{cccccc}
                \toprule
                {Method}&{mAP}&{rank-1}&{rank-5}&{rank-10}\cr
                \midrule
                Spindle~\cite{Zhao2017Spindle}&-&76.9&91.5&94.6\cr
                SVDNet~\cite{Sun2017SVDNet}&62.1&82.3&92.3&95.2\cr
                PDC~\cite{SuPose}&63.4&84.1&92.7&94.9\cr
                PSE~\cite{Sarfraz2017A}&69.0&87.7&94.5&96.8\cr
                Cam-style~\cite{Zhong2017Camera}&71.6&89.5&-&-\cr
                GLAD~\cite{WeiGLAD}&73.9&89.9&-&-\cr
                HA-CNN~\cite{LiHarmonious}&75.7&91.2&-&-\cr
                CNN-Base~\cite{fu2018towards}&79.8&92.5&-&-\cr
                PCB+RPP~\cite{Sun2017Beyond}&81.6&93.8&97.5&98.5\cr
                HPM~\cite{FuHorizontal}&82.7&94.2&97.5&98.5\cr
                SGGNN~\cite{ShenPerson}&82.8&92.3&96.1&97.4\cr
                SPReID~\cite{Kalayeh2018Human}&83.4&93.7&97.6&98.4\cr
                MHN~\cite{ChenMixed}&85.0&95.1&98.1&98.9\cr
                DG-Net~\cite{ZhengJoint}&86.0&94.8&-&-\cr
                MGN~\cite{Wang2018Learning}&86.9&95.7&-&-\cr
                DSA-ReID~\cite{ZhangDensely}&87.6&95.7&-&-\cr
                Pyramid~\cite{ZhengPyramidal}&88.2&95.7&98.4&99.0\cr
                \midrule
				{\bf CAN(ours)} &{\bf 90.6}&{\bf 96.4}&{\bf 98.8}&{\bf 99.3}\cr
                \bottomrule
            \end{tabular}
        \caption{Comparing with current state-of-the-art methods on Market-1501.}\smallskip \label{table5}
        \end{threeparttable}
    }
\end{table}


\subsection{Comparison with the State-of-the-art Methods}

In this section, we compare the results on our proposed method with existing state-of-the-art image-based person ReID methods on the dataset of Market-1501 and use DukeMTMC-ReID and CUHK03 as extended datasets.
Our experimental results are based on model accuracy without re-ranking.
Table~\ref{table5} shows the comparison with existing image-based person ReID on Market-1501 and Table~\ref{table6} shows the comparison results on DukeMTMC-ReID and CUHK03 datasets.

As we can see from Table~\ref{table5}, the existing state-of-the-art method is a pyramidal approach proposed in \cite{ZhengPyramidal} whose accuracy is mAP/rank-1=88.2\%/95.7\%, but our proposed method achieves mAP/rank-1=90.6\%/96.4\% which exceeding the previous best performance by 2.4\% in mAP and 0.7\% in rank-1.

\begin{table}[t]
	\centering
	\resizebox{\columnwidth}{!}{
		\begin{threeparttable}
			\begin{tabular}{ccccc}
				\toprule
				\multirow{2}{*}{Method}&
				\multicolumn{2}{c}{DukeMTMC-ReID}&\multicolumn{2}{c}{CUHK03} \cr
				\cmidrule(lr){2-3} \cmidrule(lr){4-5}
				&{mAP}&{rank-1}&{mAP}&{rank-1}\cr
				\midrule
                CNN-Base~\cite{fu2018towards}&68.5&83.5&59.0&63.5\cr
                PSE+ECN~\cite{Sarfraz2017A}&62.0&79.8&30.2&27.3\cr
                Cam-style~\cite{Zhong2017Camera}&57.6&78.3&-&-\cr
                GLAD~\cite{WeiGLAD}&-&-&-&85.0\cr
                HA-CNN~\cite{LiHarmonious}&63.8&80.5&41.0&44.4\cr
                PCB+RPP~\cite{Sun2017Beyond}&69.2&83.3&57.5&63.7\cr
                HPM~\cite{FuHorizontal}&74.3&86.6&63.9&57.2\cr
                MHN~\cite{ChenMixed}&77.2&89.1&65.4&77.2\cr
                DG-Net~\cite{ZhengJoint}&74.8&86.6&-&-\cr
                MGN~\cite{Wang2018Learning}&78.4&88.7&67.4&68.0\cr
                DSA-ReID~\cite{ZhangDensely}&74.3&86.2&75.2&78.9\cr
                Pyramid~\cite{ZhengPyramidal}&79.0&89.0&76.9&78.9\cr
				\midrule
				{\bf CAN(ours)} &{\bf 82.3}&{\bf 91.6}&{\bf 82.8}&{\bf 86.1}\cr
				\bottomrule
			\end{tabular}
		\caption{The comparison with existing image-based Re-ID approaches on DukeMTMC-ReID and CUHK03.}\smallskip \label{table6}
		\end{threeparttable}
	}
\end{table}

From the Table~\ref{table6}, we can notice that the pyramidal method still achieved the best performance among all previous approaches with mAP 82.3\% and rank-1 91.6\% on DukeMTMC-ReID.
In CUHK03 datasets, bouding box images are both artificially labeled and detected by detectors.
Some of the previous works were tested on labeled and detected images respectively and we select the better performance in this table.
In our experimental settings for CUHK03, all testing images are used without splitting into labeled and detected groups.
As what is shown in this table, our method achieves mAP 82.8\% and rank-1 86.1\% in model accuracy, which also outperforms all published methods by a large margin.
Our proposed Collaborative Attention Network achieves state-of-the-art performance which indicates the robustness of our method.

\section{Conclusion}

In this paper, we novelly propose the multi-branch Collaborative Attention Network (CAN), a feature extractor for person ReID task.
We find out the best combination of numbers of parts sliced from global feature maps and conclude that partition global features into too many will conversely reduce the model accuracy.
Unlike other existing part-based methods, we concatenate local features by collaborative attention mechanism to form more effective feature representations.
Our modification on loss functions in which both global and local branches are imported into Triplet Loss and Center Loss.
Through comparative experiments, we address that our proposed approach outperforms the existing state-of-the-art methods on the public datasets.

\clearpage  
{\small
\bibliographystyle{ieee_fullname}
\bibliography{egbib}
}

\end{document}